\documentclass[journal]{IEEEtran}
\usepackage{times}
\usepackage{helvet}
\usepackage{courier}
\usepackage{times}
\usepackage[dvips]{graphicx}
\usepackage{multirow}
\usepackage{epstopdf}
\usepackage{graphicx}
\usepackage{subfigure}

\usepackage{times}
\usepackage{helvet}
\usepackage{courier}
\usepackage{url}
\usepackage{array}
\usepackage{ifthen}
\usepackage{balance}
\usepackage{bm}
\usepackage{amsmath}
\usepackage{amssymb}
\usepackage{mathrsfs}
\usepackage{color}
\usepackage{graphicx}
\usepackage{algorithm,algorithmic}
\usepackage{sidecap}
\usepackage{float}
\usepackage{graphicx} %use graph format
\usepackage{epstopdf}
\usepackage{threeparttable}
\usepackage{color, colortbl}
\definecolor{Gray}{gray}{0.9}

\def\R{\mathbf{R}}
\def\X{\mathbf{X}}
\def\x{\mathbf{x}}
\def\Z{\mathbf{Z}}
\def\Zhat{\mathbf{\hat{Z}}}
\def\Ztrain{\mathbf{{\tilde{Z}}}_{tr}}
\def\Gtrain{\mathbf{{\tilde{G}}}_{tr}}
\def\z{\mathbf{\tilde{z}}}
\def\A{\mathbf{X}}
\def\P{\mathbf{P}}
\def\C{{C}}

\def\S{{K}}
\def\I{\mathbf{I}}
\def\IND{D}

\def\E{\mathbf{E}}
\def\J{\mathbf{J}}
\def\Y{\mathbf{\Gamma}}
\def\y{y}
\def\W{\mathbf{W}}
\def\Q{\mathbf{Q}}
\def\Qhide{\mathbf{\tilde{Q}}}
\def\R{\mathbf{R}}
\def\D{\mathbf{D}}
\def\q{q}

\def\D{\mathbf{D}}

\def\U{\mathbf{U}}
\def\V{\mathbf{V}}
\def\G{\mathbf{G}}
\def\Tr{\rm{Tr}}
\def\proposed{SLAP}

\definecolor{blue}{RGB}{0,0,0} %{0,0,255}
\definecolor{blue2}{RGB}{0,0,255}

\definecolor{green}{RGB}{0,0,0}%{0,100,0}

\def\figsize{0.43}
\def\figsizesmall{0.12}
\def\heightfigures{0.04}
\def\heightfigure{0.095}
\ifCLASSINFOpdf
\else
\fi
\begin{document}

\title{ Superpixelwise Low-rank Approximation based Partial Label Learning for Hyperspectral Image Classification}

\author{Shujun~Yang,~\IEEEmembership{Member,~IEEE,}
  Yu~Zhang,~\IEEEmembership{Member,~IEEE,}
  Yao~Ding,
  Danfeng~Hong,~\IEEEmembership{Senior Member,~IEEE,}

  %Yuheng~Jia,~\IEEEmembership{Member,~IEEE,}
  %  and Weijia~Zhang,~\IEEEmembership{Member,~IEEE}
%\thanks{This work was supported by }
\thanks{S. Yang is with the Guangdong Provincial Key Laboratory of Brain-inspired Intelligent Computation, Department of Computer Science and Engineering, Southern University of Science and Technology, Shenzhen 518055, China; (e-mail: yangsj3@sustech.edu.cn).}
\thanks{Y. Zhang is with the Guangdong Provincial Key Laboratory of Brain-inspired Intelligent Computation, Department of Computer Science and Engineering, Southern University of Science and Technology and also with Peng Cheng Laboratory, Shenzhen 518055, China; (e-mail: yu.zhang.ust@gmail.com).}
%\thanks{Y. Jia is with the School of Computer Science and Engineering, Southeast University, Nanjing 210096, China, and also with Key Laboratory of Computer Network and Information Integration (Southeast University), Ministry of Education, China; (e-mail: yhjia@seu.edu.cn).}
%\thanks{** is with the School of Computer Science and Engineering, ** University, ** ******, ****, and also with Key Laboratory of Computer Network and Information Integration (** University), Ministry of Education, *****; (e-mail: **@).}
\thanks{Y. Ding is with the School of Optical Engineering, Xi’an Research Institute of High Technology, Xi’an 710025, China; (e-mail: dingyao.88@outlook.com).}
\thanks{D. Hong is with the Key Laboratory of Digital Earth Science, Aerospace Information Research Institute, Chinese Academy of Sciences, Beijing 100094, China; (e-mail: hongdf@aircas.ac.cn).}
%\thanks{W. Zhang is with the School of Computer Science and Engineering, Southeast University, Nanjing 210096, China; (e-mail: zhangwj@seu.edu.cn).}
\thanks{Corresponding author: Yu Zhang}
}
\maketitle
\begin{abstract}
%The previous hyperspectral image (HSI) data contains two categories of ambiguity: 
%feature ambiguity (i.e., spectral variations) and label ambiguity (i.e., incorrect labeling or ambiguous labeling).The ambiguous labeling is incorrect labeling (e.g. noisy label) 
Insufficient prior knowledge of a captured hyperspectral image (HSI) scene may lead the experts or the automatic labeling systems to offer incorrect labels or ambiguous labels (i.e.,  assigning each training sample to a group of candidate labels, among which only one of them is valid; this is also known as partial label learning) during the labeling process. 
Accordingly, how to learn from such data with ambiguous labels is a problem of great practical importance. In this paper, we propose a novel superpixelwise low-rank approximation (LRA)-based partial label learning method, namely \proposed, which is the first to take into account partial label learning in HSI classification. 
\proposed~is mainly composed of two phases:
disambiguating the training labels and acquiring the predictive model. Specifically, in the first phase, we propose a superpixelwise LRA-based model, preparing the affinity graph for the subsequent label propagation process while extracting the discriminative representation to enhance the following classification task of the second phase. Then to disambiguate the training labels, label propagation propagates the labeling information via the affinity graph of training pixels. In the second phase, we take advantage of the resulting disambiguated training labels and the discriminative representations to enhance the classification performance. The extensive experiments validate the advantage of the proposed \proposed~method over state-of-the-art methods. 
%Finally, the resulting disambiguated training labels with the discriminative presentation are input into a typical classifier.
\end{abstract}
\IEEEpeerreviewmaketitle
\section{Introduction}
{
%In the past decade, the remote sensing community has introduced intensive works to establish an accurate hyperspectral image classifier.
%% 简介hsi 分类
The hyperspectral image (HSI) data inevitably contains two categories of ambiguity: feature ambiguity and label ambiguity. The feature ambiguity (or called the spectral variations) always exists in the acquired HSIs, i.e., pixels of the same material may change greatly, which will degrade the performance of the succeeding HSI classification task. 
To deal with the feature ambiguity arising from the noise related to complicated environmental circumstances, %(e.g., illumination, environmental, atmospheric, and temperature conditions and sensor interference), 
many low-rank approximation (LRA)-based methods \cite{mei2018simultaneous,sun2019lateral,wang2021tensor,chang2020hyperspectral} and \textcolor{blue}{other feature extraction methods \cite{liu2019review,zhang2020improved,kang2013feature}} have been proposed to relieve its impact and accordingly improve the classification performance. 
%Moreover, some other methods \cite{liu2019review,zhang2020improved,kang2013feature} have also been proposed to  issue in HSI data.
%Given the noise associated with complex environmental conditions, %%(e.g., illumination, environmental, atmospheric, and temperature conditions and sensor interference), 
%the acquired HSIs often suffer from the feature ambiguity (or called the spectral variations), i.e., pixels of the same material may change greatly which will degrade the performance of the succeeding HSI classification task. Many low-rank approximation (LRA)-based methods \cite{mei2018simultaneous,sun2019lateral,wang2021tensor,chang2020hyperspectral} have been proposed to alleviate the impact of such feature ambiguity (or called the spectral variations) and accordingly improve the classification performance. 
On the other hand, most of the existing HSI classification methods often assume that the training samples are correctly labeled. However, such an assumption may not always be true due to the requirement of extensive knowledge while labeling a captured scene.
Label ambiguity always exists, resulting from incorrect labeling or ambiguous labeling in the labeling process. 
%as the insufficient prior knowledge of a captured scene may lead to label ambiguity (e.g., incorrect labeling or ambiguous labeling) in the labeling process.
In recent years, incorrect labeling (or called noisy label learning) has already gained attention in the HSI classification task \cite{xu2021dual,tu2020hyperspectral}. 
However, ambiguous labeling (or called partial label learning \cite{wang2019partial,wang2021adaptive}) has never been considered in HSI classification, which indicates that each training instance is associated with a candidate label set, among which only one label is valid. Such a problem has been quite popular in many real-world scenarios \cite{wang2021adaptive}, e.g., computer vision and natural language processing, etc. \textcolor{blue}{In these fields, many partial label learning approaches have been proposed, in which the major strategy is trying to solve the learning task by disambiguating the candidate label set, such as graph-based disambiguation \cite{wang2019partial,wang2021adaptive,zhang2015solving}, and semi-supervised disambiguation \cite{wang2019partial}, etc}.
Partial label learning should be as pervasive as noisy label learning in HSI classification due to the following reasons:

1) The land cover is very complex, e.g., low interclass discriminability and high intraclass variations, and the information offered to the expert or the automatic
labeling system is very limited. 
%%空间不够时候；下句可以省略
%They may find it difficult to distinguish some specific classes which look similar to each other; 
%%%%%空间不够时候这句可以省略
They may find it challenging to tell the ground-truth labels of some confusing pixels whose class labels seem ambiguous, i.e., several classes all seem likely to be the ground-truth class associated with such a confusing pixel. %Hence, offering a candidate label set is a more natural way to address such a case than only one ground-truth label.
Hence, a natural way to address such a case is to provide a candidate label set instead of one ground-truth label.

%if multiple experts label the same image at the same time, the labeling results may be inconsistent with different experts 
2) If multiple experts are invited to label the pixels of the same land cover at the same time, the labeling results for the same pixel may be inconsistent among different experts, accordingly collecting all candidate labels marked by them for an identical pixel seems a commonly used way in such a labeling system. 

% 现有的partial label learning 方法 没有考虑到高光谱图片的spatial 信息， 没有考虑到HSI 的feature ambiguity 鉴于此 我们提出一个方法； 介绍下方法
However, the previous HSI classification methods never considered the above partial label learning problem \cite{wang2019partial,wang2021adaptive}, which has been well studied. Moreover, existing partial label learning methods are insufficient to capture the data characteristics of an HSI, such as the \textcolor{blue}{local spatial structure} \textcolor{blue}{\cite{xu2015spectral,fan2017hyperspectral}}, i.e., the pixels within a small local region often come from the same class.
Accordingly, we propose a novel superpixelwise LRA-based partial label learning method, in which the data characteristics inherited in an HSI and partial label learning are combined into one unified framework. Precisely, such a framework mainly consists of two phases, i.e., disambiguating the training labels and acquiring the predictive model. In the first phase, we build a superpixelwise LRA-based model, which generates the local affinity graph for the subsequent label propagation process and extracts the discriminative representation alleviating the feature ambiguity. After that, the resulting affinity graph is then used to assist the label propagation process, which can disambiguate the training labels. After that, in the second phase, the resulting disambiguated labels of training data and the discriminative representations of all pixels are then input into a typical classifier. Experiments demonstrate that the proposed \proposed~method outperforms state-of-the-art methods significantly. 

%The rest of this paper is organized as follows. In Section \ref{sec:method}, we present the proposed \proposed~method, followed by the extensive experiments and discussions in Section \ref{sec:expriments}. Finally, Section \ref{sec:conclusion} concludes this paper. 

%Conventional supervised learning often assumes that each training instance is associated with a ground-truth label. However, in many real-world applications, one can only get access to a candidate label set associated with each training instance among which only one label is valid.
%Label noise has also received attention in HSI in recent years \cite{xu2021dual,jiang2018hyperspectral,tu2020hyperspectral}, which they 

%% label的ambigulity weakly supervised learning method 在HSI里很少有人考虑，现有的基本考虑noisy label, partial label  很少有人考虑，介绍下重要性

%% feature 的ambigulity 另一方面数据特征的噪声对于分类学习也会带来负面影响，目前有些工作考虑LRA-based的噪声， 

%% 我们工作同时考虑两个方面， 即 Label的ambiguity 和 feature 的ambigulity 创新点
}

\section{Methodology}\label{sec:method}
Let $\mathcal{Y}=[\y_1,\y_2,\ldots,\y_c]$ be the label space concerning $c$ class labels. Partial label learning targets to acquire a predictive model from the given training data 
$\mathbb{D}=\{(\x_i,\C_i)| 1 \leq i \leq p\}$, where $\x_i\in \mathbb{R}^{d \times 1}$ denotes the feature representation of the $i$-th training pixel, i.e., a $d$-dimensional feature vector; and $\C_i \subseteq \mathcal{Y}$ denotes its candidate label set.
%is the $i$-th training pixel, and $\C_i \subseteq \mathcal{Y}$ represents the corresponding candidate label set. 

The proposed \proposed~method consists of two phases: disambiguating the training labels and acquiring the predictive model.
In the first phase, we propose a Laplacian regularized local LRA-based model and conduct it on each homogeneous region, i.e., superpixel. Such a strategy is capable of generating the local affinity graph for the succeeding label propagation process while extracting the discriminative representation to enhance the subsequent classification process (see more details in Section \ref{sec:part1}). After that, to disambiguate the training label, we build the similarity graph corresponding to the training data and perform the label propagation (see more details in Section \ref{sec:part2}). In the second phase, the resulting discriminative representation combined with the disambiguated train labels are input into a specific classifier, such as SVM (see more details in Section \ref{sec:predict}).
In the following, we introduce the proposed \proposed~in detail.
\subsection{Local Affinity Graph and Discriminative Representation}\label{sec:part1}
After performing the superpixel segmentation (i.e., the entropy rate superpixel method \cite{liu2011entropy}) on an HSI, the HSI image $\X$ is segmented into $\S$ superpixels, i.e., [$\X_1$, $\X_2$, \ldots, $\X_{\S}$], where $\X_i \in \mathbb{R}^{d\times n_i}$ is the $i$-th superpixel containing $n_i$ pixels.
For each superpixel, we formulate a Laplacian regularized local LRA-based model as follows, which can output the discriminative feature representation that removes the noises (i.e., $\X_i\Z_i$ in problem (\ref{eq:obj1})), and the local affinity graph (i.e., the coefficient matrix $\Z_i$) for the succeeding label propagation process (see more details in the succeeding Section \ref{sec:part2}).
\subsubsection{The Model}\label{sec:model}
\begin{eqnarray}\label{eq:obj1}\small
\begin{aligned}
\min_{\Z_i,\E_i} &\|\Z_i\|_*+ \lambda \|\E_i\|_{2,1}+\gamma \Tr(\A_i\Z_i\G_i(\A_i\Z_i)^T)\\
s.t. & \X_i=\A_i\Z_i+\E_i,\Z_i\geq 0,
\end{aligned}
\end{eqnarray}
where $\|\cdot\|_{2,1}$ and $\|\cdot\|_*$ are the $\ell_{2,1}$ norm and nuclear norm operations, respectively, $\lambda$ and $\gamma$ are the trade-off parameters, $\E_i$, $\Z_i$, and $\G_i$ are the noisy part, the coefficient matrix, and the Laplacian matrix corresponding to $\X_i$, respectively. Precisely, we follow the suggestions in \cite{yang2022local} to build the superpixel-guided Laplacian matrix $\G_i$ as a prior.

%In problem (\ref{eq:obj1}), the first term (i.e., $\|\Z_i\|_*$) perform the low-rank approximation on the coefficient matrix $\Z_i$. The second term (i.e., $\lambda \|\E_i\|_{2,1}$) makes the noisy part $\E_i$ sparse by using $\ell_{2,1}$ norm. The third term embeds the manifold structure of the graph $\G_i$ into the denoised feature representation $\A_i\Z_i$.

\subsubsection{Optimization}
To solve problem (\ref{eq:obj1}), we adopt two auxiliary variables $\W_i$ and $\J_i$ to reformulate it as 
\begin{eqnarray}\label{eq:objaug}\small
\begin{aligned}
\min_{\W_i,\Z_i,\E_i,\J_i}&\|\W_i\|_*+\lambda \|\E_i\|_{2,1}+\gamma \Tr(\A_i\J_i\G_i\J_i^{T}\A_i^{T})\\
s.t.\ & \X_i=\A_i\Z_i+\E_i,\Z_i \geq 0, \Z_i=\J_i,\W_i=\Z_i.
\end{aligned}
\end{eqnarray}
The inexact augmented Lagrangian multiplier (IALM) \cite{lin2010augmented}
is employed to solve problem (\ref{eq:objaug}), which alternatively updates the variables $\W_i$, $\Z_i$, $\E_i$, and $\J_i$. %by solving the subproblems respectively in each iteration.
The augmented Lagrangian of problem (\ref{eq:objaug}) can be written as
%\begin{eqnarray}
\begin{align}\small
\min_{\W_i,\Z_i,\E_i,\J_i}&\|\W_i\|_*+\lambda \|\E_i\|_{2,1}+\gamma \Tr(\A_i\J_i\G_i\J_i^{T}\A_i^{T})\notag\\
&+\frac{\mu}{2}\|\X_i-\A_i\Z_i-\E_i+\frac{\Y_{1,i}}{\mu}\|_{F}^2\\
&+\frac{\mu}{2}\|\Z_i-\J_i+\frac{\Y_{2,i}}{\mu}\|_{F}^2
+\frac{\mu}{2}\|\Z_i-\W_i+\frac{\Y_{3,i}}{\mu}\|_{F}^2\notag\\
s.t.\ & \Z_i \geq 0, \notag
\end{align}
%\end{eqnarray}
where %$\Y_{1,i} \in \mathbb{R}^{d\times n_i}$,$\Y_{2,i},\Y_{3,i} \in \mathbb{R}^{n_i \times n_i}$ 
$\Y_{1,i},\Y_{2,i},\Y_{3,i}$
are three Lagrange multipliers, $\|\cdot\|_F$ denotes the Frobenius norm, and $\mu$ is a positive scalar. %The four subproblems are updated as follows. 
In the following part, we show how to update the four subproblems.

a) By keeping other variables fixed, the $\W_i$ subproblem becomes
\begin{eqnarray}\label{eq:w}\small
\begin{aligned}
\min_{\W_i}\|\W_i\|_{*}+\frac{\mu}{2}\|\P_i-\W_i\|_F^2,
\end{aligned}
\end{eqnarray}
 where $\P_i=\Z_i+\frac{\Y_{3,i}}{\mu}$. 
Let $\U_i\mathbf{\Sigma}_{i}\V_i^\mathsf{T}$ be the singular value decomposition (SVD) of $\P_i$,
%where $\mathbf{\Sigma}_{i}$ is a diagonal matrix with the singular values on its diagonal, and the columns of $\U_i$ and $\mathbf{V}_i$ are the left and right singular vectors, respectively. 
%Problem (\ref{eq:w}) has
the closed-form solution %\cite{cai2010singular} 
of problem (\ref{eq:w}) is obtained as
\begin{eqnarray}\label{eq:w_s}\small
\begin{aligned}
\W_i=\U_i\mathcal{T}_{\mu^{-1}}\left (\mathbf{\Sigma}_{i}\right)\V_i^\mathsf{T},
\end{aligned}
\end{eqnarray}
where $\mathcal{T}_{\varepsilon}(w)$ is a shrinkage operator, i.e., $\mathcal{T}_{\varepsilon}(w):={\rm  sgn}(w){(|w|-\varepsilon)}_{+}$, and ${\rm  sgn}(\cdot)$ represents the signum function indicating the sign of a number.
%i.e., $\mathcal{S}_{\varepsilon}(x)=0$ if $|x|\leq \varepsilon$ and $\mathcal{S}_{\varepsilon}(x)=\left(1-\varepsilon/|x|\right)x$ otherwise.

b) By keeping other variables fixed, the $\Z_i$ subproblem becomes
\begin{eqnarray}\label{eq:z}\small
\begin{aligned}
\min_{\Z_i\geq 0}&\frac{\mu}{2}\|\X_i-\A_i\Z_i-\E_i+\frac{\Y_{1,i}}{\mu}\|_F^2\\
&+\frac{\mu}{2}\|\Z_i-\J_i+\frac{\Y_{2,i}}{\mu}\|_F^2+\frac{\mu}{2}\|\Z_i-\W_i+\frac{\Y_{3,i}}{\mu}\|_F^2.\\
%s.t. &\Z_i \geq 0
\end{aligned}
\end{eqnarray}
Problem (\ref{eq:z}) has a closed-form solution as
\begin{equation}\label{eq:z_s1}\small
\Z_i=\max\left(\Zhat_i,0 \right),
\end{equation}where
$\Zhat_i$ is formulated as 
%$\Zhat_i=(\A_i^{T}\A_i+2\I)^{-1}\left [\A_i^{T}(\X_i-\E_i+\frac{\Y_{1,i}}{\mu})+\J_i-\frac{\Y_{2,i}}{\mu}+\W_i-\frac{\Y_{3,i}}{\mu} \right ]$ 
\begin{eqnarray}\label{eq:z_s}\small
\begin{aligned}
\Zhat_i=&{\left(\A_i^{T}\A_i+2\I\right)}^{-1} \left( \A_i^{T}\X_i-\A_i^{T}\E_i+\frac{\A_i^{T}\Y_{1,i}}{\mu}\right.\\
&+\left.{\J_i-\frac{\Y_{2,i}}{\mu}+\W_i-\frac{\Y_{3,i}}{\mu}}\right),
\end{aligned}
\end{eqnarray}
where $\mathbf{I}$ is an identity matrix.

c) By keeping other variables fixed, the $\E_i$ subproblem becomes
\begin{eqnarray}\label{eq:e}\small
\begin{aligned}
\min_{\E_i} &\lambda \|\E_i\|_{2,1}+\frac{\mu}{2}\|\X_i-\A_i\Z_i-\E_i+\frac{\Y_{1,i}}{\mu}\|_F^2.
\end{aligned}
\end{eqnarray}

Problem (\ref{eq:e}) has a closed-form solution \cite{liu2010robust} as 
\begin{eqnarray}\label{eq:e_s}\small
\begin{aligned}
%\E_i=\mathcal{\tilde{T}}_{\lambda/\mu}\left ( \X_i-\A_i\Z_i+\frac{\Y_{1,i}}{\mu}\right),
\E_i(:,j)=&\left\{\begin{matrix}
 \frac{\|\D_i(:,j)\|-\frac{\lambda}{\mu}}{\|\D_i(:,j)\|}\D_i(:,j),& \textrm{if}\,\frac{\lambda}{\mu} < \|\D_i(:,j)\|,\\ 
0, & \textrm{otherwise},
\end{matrix}\right.\\
&j=1,2,\ldots,n_i.
\end{aligned}
\end{eqnarray}
where $\D_i=\X_i-\A_i\Z_i+\frac{\Y_{1,i}}{\mu}$,
$\E_i(:,j)$ and $\D_i(:,j)$ represent the $j$th column of $\E_i$ and $\D_i$, respectively.

d) By keeping other variables fixed, the $\J_i$ subproblem becomes
\begin{eqnarray}\label{eq:J}\small
\begin{aligned}
\min_{\J_i}\gamma \Tr\left (\A_i\J_i\G_i\J_i^{T}\A_i^{T}\right )+\frac{\mu}{2}
\|\Z_i-\J_i+\frac{\Y_{2,i}}{\mu}\|_F^2.
\end{aligned}
\end{eqnarray}
By taking the derivative of problem (\ref{eq:J}) with respect to $\J_i$ and setting it to zero, we have
\begin{eqnarray}\label{eq:J_s}\small
\begin{aligned}
2\gamma\A_i^{T}\A_i\J_i+\mu\J_i\G_i^{\dagger}=\mu\left(\Z_i+\frac{\Y_{2,i}}{\mu}\right)\G_i^{\dagger},
\end{aligned}
\end{eqnarray}
where $\G_i^{\dagger}$ denotes the pseudoinverse matrix of $\G_i$.
Problem (\ref{eq:J_s}) is a Sylvester equation, which can be addressed by using the MATLAB function $sylvester\left( \right)$.

e) The Lagrange multipliers $\Y_{1,i}$, $\Y_{2,i}$, $\Y_{3,i}$ and the positive scalar $\mu$ are updated as
\begin{align}\label{eq:lagrange}\small
&\Y_{1,i}^{\rm{iter+1}}=\Y_{1,i}^{\rm{iter}}+\mu\left (\X_i-\A_i\Z_i-\E_i\right),\nonumber\\
&\Y_{2,i}^{\rm{iter+1}}=\Y_{2,i}^{\rm{iter}}+\mu\left ( \Z_i-\J_i\right ),\\
&\Y_{3,i}^{\rm{iter+1}}=\Y_{3,i}^{\rm{iter}}+\mu\left ( \Z_i-\W_i\right ),\nonumber\\
&\mu^{\rm{iter+1}}=\min \left(\mu_{\max},\rho\mu^{\rm{iter}}\right),\nonumber
\end{align}
where iter indicates the iteration index, and the parameter $\rho>1$ increases the speed of convergence. 

Algorithm 1 summarizes the IALM-based optimization algorithm for solving the LRA-based model of the proposed \proposed~method in problem (\ref{eq:obj1}). %which shows the initial values of the variables and the convergence condition.
%$\mu_{\max}=10^{12}$, and $\mu$ is initialized as $10^{-4}$.
%$\rho=1.1$

% The convergence condition for the whole algorithm is $\|\X_i-\A_i\Z_i-\E_i\|_{\infty} \le \epsilon$, $\|\Z_i-\J_i\|_{\infty} \le \epsilon$, and $\|\Z_i-\W_i\|_{\infty} \le \epsilon$ where $\epsilon=10^{-3}$ and $\|\cdot\|_{\infty}$ is the max norm of a matrix. In addition, the parameter $\mu$ is initialized as $10^{-4}$ with the maximum value $\mu_{\max}=10^{12}$ and $\rho=1.1$.
\begin{algorithm}[!htbp]\small
\caption{
Algorithm for solving problem (\ref{eq:obj1}).
}
\begin{algorithmic}[1]
\renewcommand{\algorithmicrequire}{\textbf{Input:}}
\renewcommand{\algorithmicensure}{\textbf{Output:}}
\REQUIRE $\X_i$, $\G_i$, and regularization parameters $\lambda$, $\gamma$.
\ENSURE  $\Z_i$, $\E_i$.
\STATE \textbf{Initialize:} %$\W_i=\Z_i=\E_i=\J_i=\Y_{1,i}=\Y_{2,i}=\Y_{3,i}=\bf{0}$, 
$\W_i \leftarrow \bf{0}$, $\Z_i \leftarrow \bf{0}$, $\E_i \leftarrow \bf{0}$, $\J_i \leftarrow \bf{0}$, $\Y_{1,i} \leftarrow \bf{0}$,
$\Y_{2,i} \leftarrow \bf{0}$, $\Y_{3,i} \leftarrow \bf{0}$,
$\mu=10^{-4}$, $\mu_{\max}=10^{12}$, $\epsilon=10^{-3}$, $\rho=1.1$;
%\WHILE{$\|\X-\LL-\E\|_{\infty} > \epsilon$ or $\|\LL-\J\|_{\infty} > \epsilon$}
\WHILE{not converged}
\STATE Update $\W_i$ via the closed-form (\ref{eq:w_s});
\STATE Update $\Z_i$ via the closed-form (\ref{eq:z_s1});
\STATE Update $\E_i$ via the closed-form (\ref{eq:e_s});
\STATE Update $\J_i$ by solving (\ref{eq:J_s});
\STATE Update $\Y_{1,i}$, $\Y_{2,i}$, $\Y_{3,i}$, and $\mu$ by using (\ref{eq:lagrange});
\STATE Check the convergence condition: $\|\X_i-\A_i\Z_i-\E_i\|_{\infty} \le \epsilon$, $\|\Z_i-\J_i\|_{\infty} \le \epsilon$, and $\|\Z_i-\W_i\|_{\infty} \le \epsilon$.
\ENDWHILE
\end{algorithmic}
\end{algorithm}
\subsection{Label Propagation for Disambiguation}\label{sec:part2}
In partial label learning, label propagation has been proven to be effective \cite{wang2019partial%,lyu2020partial
} in disambiguating the training labels, which propagates the labeling information via the affinity graph and thus leads to the connection between the feature and label spaces.

%The affinity graph (i.e., the coefficient matrices $\mathcal{Z}=\{\Z_1,\Z_2,\ldots,\Z_S\}$) that output in Section \ref{sec:part1} can reflect the similarities between pixels. 
Concretely, with the resulting affinity graphs (i.e., the coefficient matrices $\mathcal{Z}=\{\Z_1,\Z_2,\ldots,\Z_{\S}\}$) that output in Section \ref{sec:part1}, we first denote $\Ztrain$ as the affinity matrix corresponding to the training data $\mathbb{D}$, which is defined as 
%that can be easily obtained from $\mathcal{Z}$
\begin{eqnarray}\small
\begin{aligned}
\Ztrain^{i,j}=\left\{\begin{matrix}
 \Z_{{\IND_i}}^{v_i,v_j},& if \ \IND_{i}=\IND_{j}, \\
 0,&  otherwise,\\
\end{matrix}\right.
\end{aligned}
\end{eqnarray}
where $\IND_{i}$ and $\IND_{j}$ respectively denote the indices of superpixels corresponding to the $i$-th and $j$-th training pixel; $v_i$ and $v_j$ are the indices in $\Z_{\IND_{i}}$ that the $i$-th and $j$-th training pixel located in, respectively.
To promote the following label propagation process, we normalize $\Ztrain$ by column, i.e., $\z_{tr}^i=\frac{\z_{tr}^i}{\|\z_{tr}^i\|_2}$, where $\z_{tr}^i$ is the $i$-th column of $\Ztrain$. Then, the affinity graph of the training data can be expressed as $\Gtrain=\frac{\Ztrain+\Ztrain^{T}}{2}$.

%Then the label propagation strategy is utilized to construct the relationship between the feature space and label space.
For the subsequent label propagation, we then denote $\Q={[\q_{i,j}]}_{p\times c}$ as the labeling confidence matrix, where $\q_{i,b}\geq 0$ that represents the probability of the label $\y_{b}$ being the ground-truth label of $\x_i$. As each training pixel owns several candidate labels in partial label learning, we equally treat each candidate label and initialize the labeling confidence matrix $\Q$ as
\begin{eqnarray}\small
\begin{aligned}
 1 \leq i \leq p: \Q_{i,b}^{(0)}=\left\{\begin{matrix}
 \frac{1}{|\C_i|},&  if\ \y_{b} \in \C_i,\\
 0,&  if\ \y_{b} \notin \C_i.\\
\end{matrix}\right.
\end{aligned}
\end{eqnarray}

 %Let $\Qhide^{(t)}$ be the labeling confidence matrix of the $t$th iteration, which can be obtained through propagating the labeling information along with the weight matrix $\G_{tr}$.
%Generally, we update the labeling confidence matrix by propagating the labeling information along with the affinity graph $\Gtrain$, which is defined as follows:
Generally, to update the labeling confidence matrix, we propagate the labeling information in conjunction with the affinity graph $\Gtrain$, which can be described as 
\begin{eqnarray}\small
\begin{aligned}
\Qhide^{(t)} = (1-\alpha)\cdot \Q^{(0)}+\alpha\cdot \Gtrain\cdot \Q^{(t-1)},
\end{aligned}
\end{eqnarray}
where $\Qhide^{(t)}$ denotes the labeling confidence matrix of the $t$-th iteration, and $\alpha\in (0,1)$ makes a trade-off between the initial labeling confidence matrix (i.e., $\Q^{(0)}$) and the outcome inheriting from the previous iteration (i.e., $\Gtrain\cdot \Q^{(t-1)}$).
We thereafter rescale $\Qhide^{(t)}$ into $\Q^{(t)}$ as follows:
\begin{eqnarray}\small
\begin{aligned}
1\leq i \leq p: \Q_{i,b}^{(t)}=\left\{\begin{matrix}
 \frac{\Qhide_{i,b}^{(t)}}{\sum_{\y_l\in \C_i}\Qhide_{i,l}^{(t)} },&  if\ \y_{b} \in \C_i,\\
 0,&  if\ \y_{b} \notin \C_i.\\
\end{matrix}\right.
\end{aligned}
\end{eqnarray}
\subsection{Predictive Model}\label{sec:predict}
After learning $\Q$ (see more details in Section \ref{sec:part2}), the disambiguated labels $\R$ of training data can be defined as 
\begin{eqnarray}\small
\begin{aligned}
\R_{i}=\arg\max_{\y_l\in \C_i}\Q_{i,l}.
\end{aligned}
\end{eqnarray}
These resulting disambiguated labels $\R$ and the discriminative representations (or called the denoised feature representations) for all $\S$ homogeneous regions (see more details in Section \ref{sec:part1}), i.e., $[\X_1\Z_1,\X_2\Z_2,\ldots,\X_{\S}\Z_{\S} ]$, are then input into a typical classifier (e.g., SVM).
\section{Experiments}\label{sec:expriments}

\subsection{Experimental Datasets and Settings}
In this section, the experiments are conducted on two commonly used benchmark datasets\footnote{\scriptsize\url{http://www.ehu.eus/ccwintco/index.php/Hyperspectral_Remote_Sensing_Scenes}}, i.e., \textit{Indian Pines} and \textit{Salinas Valley}. The \textit{Indian Pines} dataset contains an image of size $145 \times 145$ with 16 classes, where each pixel is represented in a 200-dimensional spectral band. The \textit{Salinas Valley} dataset contains an image of size $512 \times 217$ with 16 classes, where each pixel is represented in a 204-dimensional spectral band.
%\subsubsection{Indian Pines Dataset}
%This scene consists of $145 \times 145$ pixels with 200-dimensional spectral bands, which is collected by Airborne Visible and InfraRed Imaging Spectrometer (AVIRIS) sensor over Indian Pines test site.
%\textcolor{green}{
%There are 10249 pixels from 16 classes in the ground-truth map.}
%\subsubsection{Salinas Valley Dataset}
%This scene consists of $512 \times 217$ pixels with 204-dimensional spectral bands, which is acquired by the AVIRIS sensor over Salinas Valley.
%\textcolor{green}
%{There are 54129 pixels from 16 classes in the ground-truth map.}

Three common evaluation criteria are selected to assess the performance, i.e., overall accuracy (OA), average accuracy (AA), and Kappa coefficient ($\kappa$).
Concretely, following the widely used controlling protocol \cite{wang2019partial,wang2021adaptive} %lyu2020partial
in the partial label setting,
%we denote $r$ as the number of false labels in the candidate label sets (i.e., $|\C_i|=r+1$). More specifically, 
the candidate label set of each training pixel comprises the ground-truth label and $r$ randomly selected false labels from $\mathcal{Y}$ (i.e., $|\C_i|=r+1$).
%More specifically, to construct the candidate label set for each training pixel, 
%except the ground-truth label, we randomly select $r$ remaining labels from $\mathcal{Y}$.
Moreover, our experiments set the candidate values of $r$ as 1 or 2.

Following the suggestion in \cite{yang2022local}, we empirically set the number of superpixels as 64 and 150 for \textit{Indian Pines} and \textit{Salinas Valley}, respectively. 
\subsection{Parameter Analysis}
%Here we analyze the effect of the parameters $\lambda$, $\gamma$, and $\alpha$ on the classification performance of the proposed \proposed~method. 
Here we conduct the parameter analysis on the parameters $\lambda$, $\gamma$, and $\alpha$ of the proposed \proposed~method. 
Specifically, the values of $\lambda$ and $\gamma$ are selected from $\{0.01,0.1,1\}$ and $\{0, 0.001, 0.01, 0.1, 1, 2, 5, 10, 20, 50, 70, 100\}$ respectively. 

\begin{figure}[t]
\subfigure[]{
\begin{minipage}[t]{\figsize\linewidth}
\centering
\includegraphics[width=1.0\textwidth , height=\heightfigure\textheight]{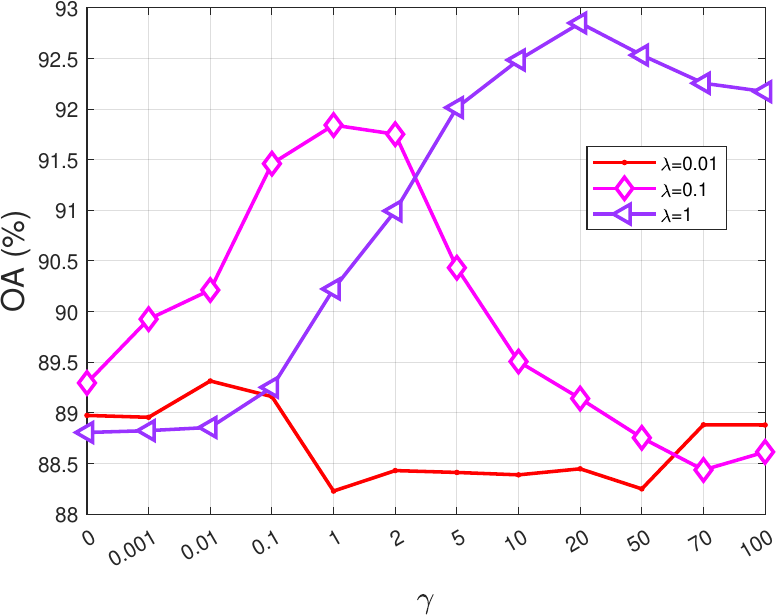}
\end{minipage}}
\centering
\subfigure[]{
\begin{minipage}[t]{\figsize\linewidth}
\centering
\includegraphics[width=1.0\textwidth , height=\heightfigure\textheight]{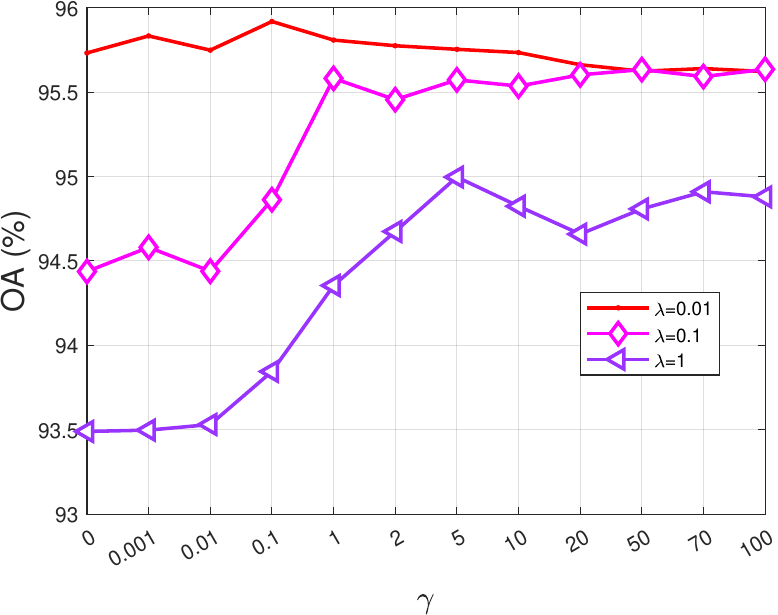}
\end{minipage}}
\caption{Illustration about how $\lambda$ and $\gamma$ affect the OA of \proposed. Here we set $r=1$ and the training percentage of each class as 5\% and 1\% for \textit{Indian Pines} (a) and \textit{Salinas Valley} (b), respectively.}
\label{fig:paras_lambda_gamma}
\end{figure}

As shown in Fig. \ref{fig:paras_lambda_gamma}, with a fixed value of $\alpha$, i.e., 0.9, we first study how the performance of \proposed~ changes with values of $\lambda$ and $\gamma$ changing. 
We can find that the highest OAs differ greatly with different values of $\lambda$, e.g., the value increases from 89.31 \% to 92.85 \%  with $\lambda$ increasing from 0.01 to 1 on \textit{Indian Pines}.
Furthermore, with a specified value of $\lambda$, the value of OA often first increases and then decreases with the increasing $\gamma$.
Moreover, the proposed \proposed~method achieves the highest OA with $\lambda=1$ and $\gamma=20$ for \textit{Indian Pines}, and $\lambda=0.01$ and $\gamma=0.1$ for \textit{Salinas Valley}.

\begin{figure}[t]
\subfigure[]{
\begin{minipage}[t]{\figsize\linewidth}
\centering
\includegraphics[width=1.0\textwidth , height=\heightfigure\textheight]{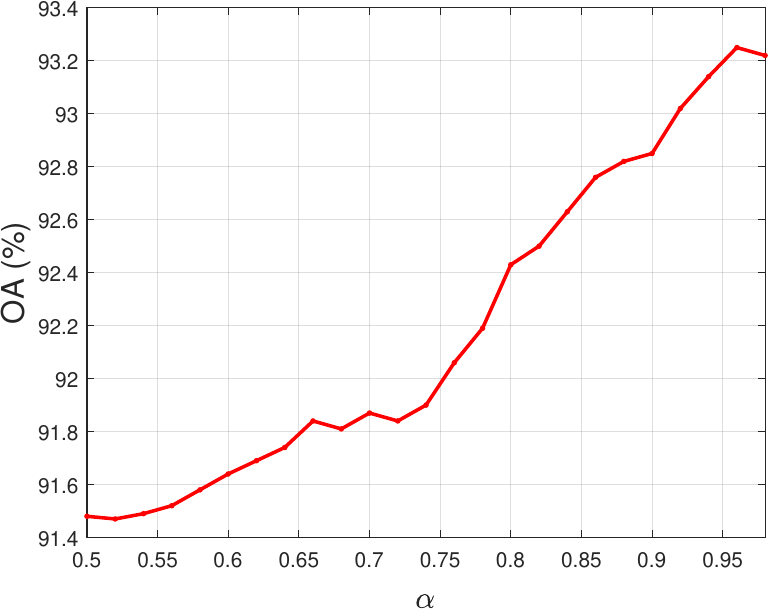}
\end{minipage}}
\centering
\subfigure[]{
\begin{minipage}[t]{\figsize\linewidth}
\centering
\includegraphics[width=1.0\textwidth , height=\heightfigure\textheight]{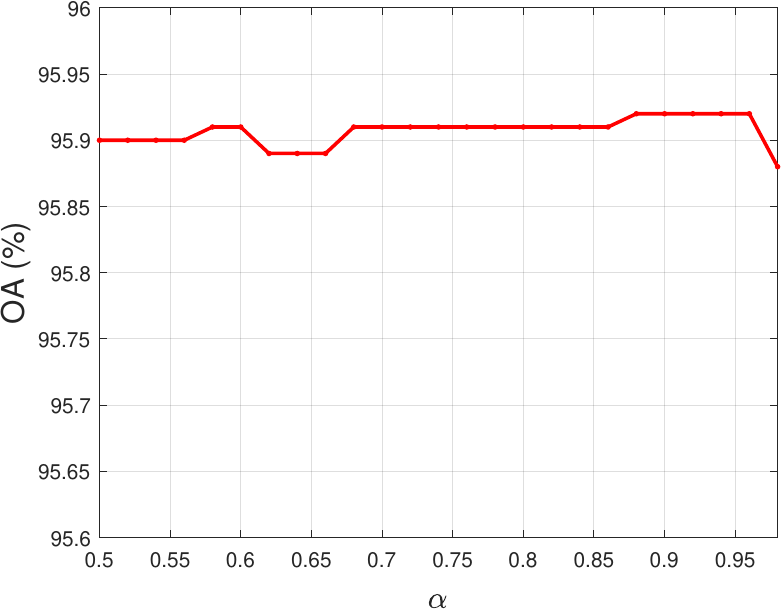}
\end{minipage}}
\caption{%Illustrate the effect of the parameter $\alpha$ on the classification performance of the proposed \proposed~method concerning OA. Here we set $r=1$ and the training percentage of each class as 5\% and 1\% for \textit{Indian Pines} and \textit{Salinas Valley}, respectively. (a) for \textit{Indian Pines} and (b) for \textit{Salinas Valley}.
Illustration about how $\alpha$ affects the OA of \proposed. Here we set $r=1$ and the training percentage of each class as 5\% and 1\% for \textit{Indian Pines} (a) and \textit{Salinas Valley} (b), respectively.}
\label{fig:alpha_oa}
\end{figure}

Furthermore, we also investigate the effect of the parameter $\alpha$. Fig. \ref{fig:alpha_oa} shows how the parameter $\alpha$ affects the performance on both datasets.

Consequently, in the succeeding experiments, we set the parameters $\lambda$ and $\gamma$ to the optimal values according to Fig. \ref{fig:paras_lambda_gamma}.
Furthermore, the parameter $\alpha$ is set as 0.96 for both datasets.

\begin{figure}[t]
\subfigure[]{
\begin{minipage}[t]{\figsize\linewidth}
\centering
\includegraphics[width=1.0\textwidth , height=\heightfigure\textheight]{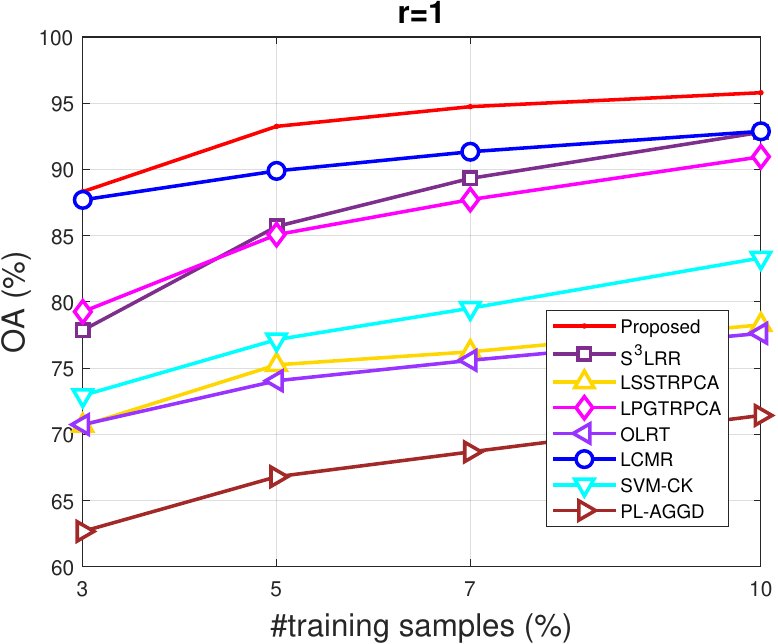}
\end{minipage}}
\centering
\subfigure[]{
\begin{minipage}[t]{\figsize\linewidth}
\centering
\includegraphics[width=1.0\textwidth , height=\heightfigure\textheight]{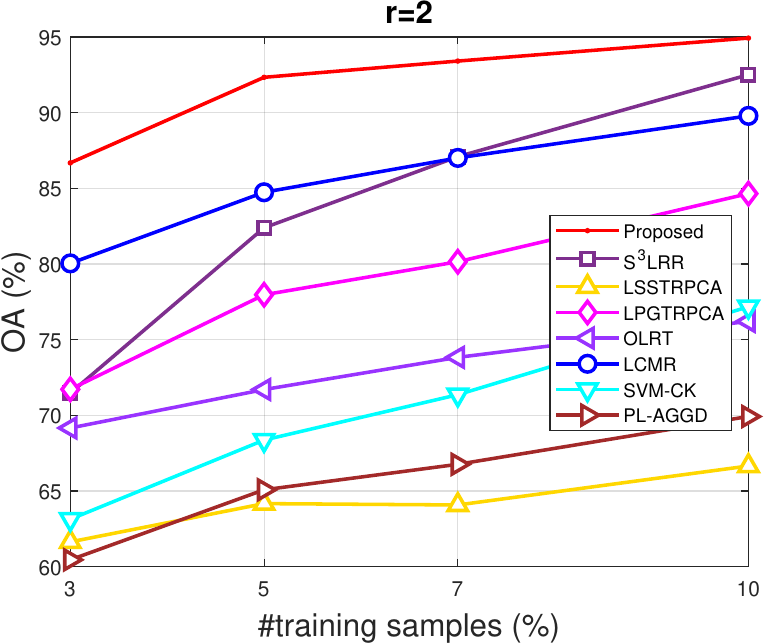}
\end{minipage}}

\subfigure[]{
\begin{minipage}[t]{\figsize\linewidth}
\centering
\includegraphics[width=1.0\textwidth , height=\heightfigure\textheight]{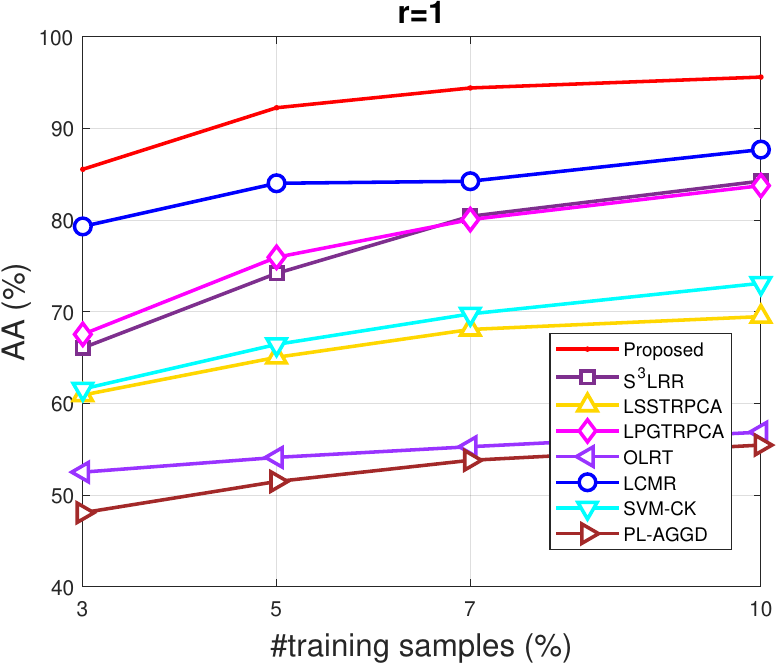}
\end{minipage}}
\centering
\subfigure[]{
\begin{minipage}[t]{\figsize\linewidth}
\centering
\includegraphics[width=1.0\textwidth , height=\heightfigure\textheight]{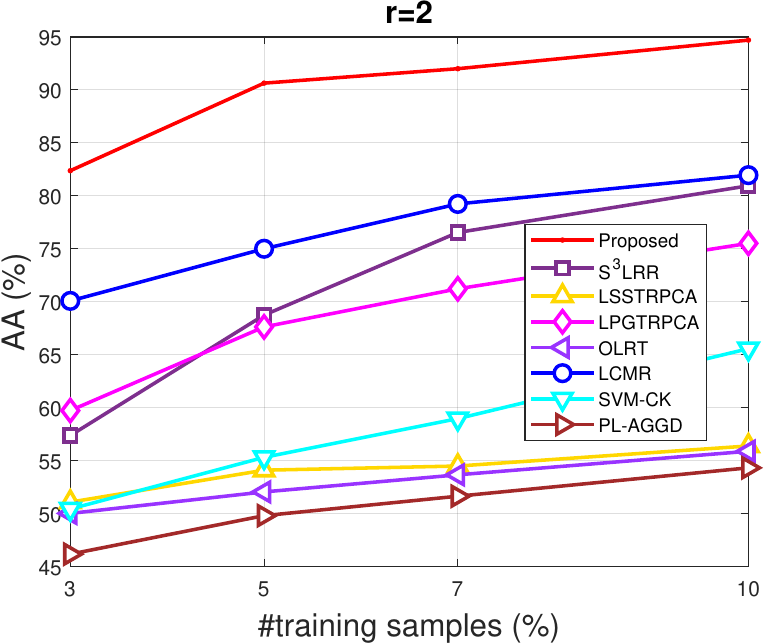}
\end{minipage}}

\subfigure[]{
\begin{minipage}[t]{\figsize\linewidth}
\centering
\includegraphics[width=1.0\textwidth , height=\heightfigure\textheight]{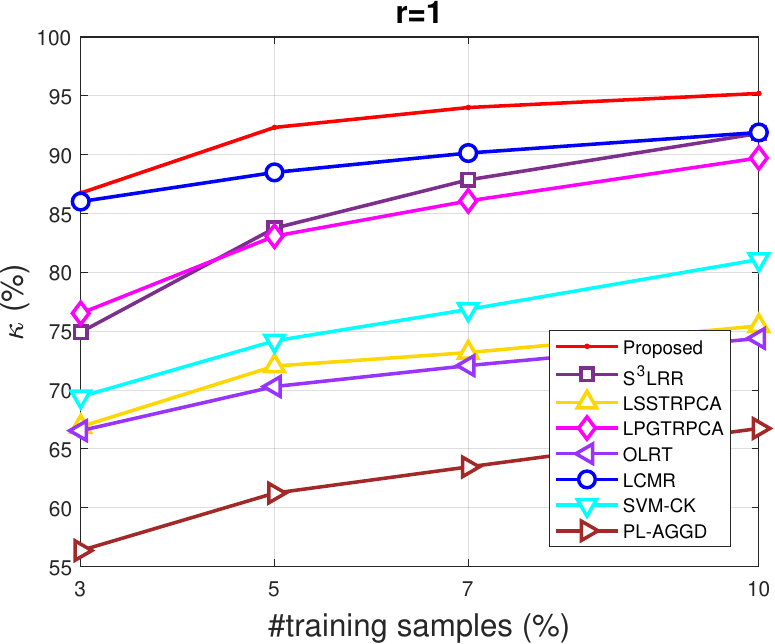}
\end{minipage}}
\centering
\subfigure[]{
\begin{minipage}[t]{\figsize\linewidth}
\centering
\includegraphics[width=1.0\textwidth , height=\heightfigure\textheight]{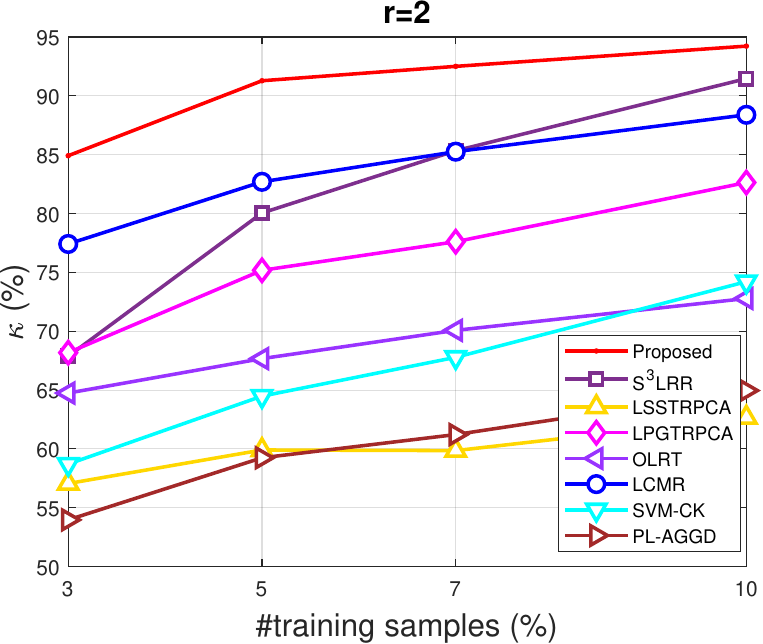}
\end{minipage}}
\caption{The performance comparisons concerning OA, AA, and $\kappa$ under various training percentages of each class with $r=1$ or 2 on \textit{Indian Pines}. Specifically, (a)-(b) for OA with $r=1$ and 2, (c)-(d) for AA with $r=1$ and 2, and (e)-(f) for $\kappa$ with $r=1$ and 2. }
\label{fig:results_indian}
\end{figure}

\subsection{Comparison of Classification Performance}
%Comparison with State-of-the-Art Methods}
%We compare the proposed \proposed~method with state-of-the-art HSI classification methods,
%including \textcolor{blue}{
%seven LRA-based methods for HSIs, i.e., RPCA \cite{candes2011robust},  LLRSSTV \cite{he2018hyperspectral}, S$^3$LRR \cite{mei2018simultaneous}, SS-LRR \cite{fan2017hyperspectral}, LSSTRPCA \cite{sun2019lateral},
 %LPGTRPCA \cite{wang2021tensor}, OLRT \cite{chang2020hyperspectral}.}

We validate the advantage of the proposed \proposed~method by comparing it with several state-of-the-art HSI classification methods, and one partial label learning method \cite{wang2021adaptive} which is not for HSI; including a) four LRA-based methods, i.e., S$^3$LRR \cite{mei2018simultaneous}, LSSTRPCA \cite{sun2019lateral}, LPGTRPCA \cite{wang2021tensor}, and OLRT \cite{chang2020hyperspectral}, b) two spatial-spectral feature extraction methods, i.e., LCMR \cite{lcmr} and SVM-CK \cite{svm-ck},
%b) one supervised method, i.e., CCJSR \cite{tu2018hyperspectral} 
and c) one partial label learning method, i.e., PL-AGGD \cite{wang2021adaptive}. Along with the recovered representation, 
all the LRA-based methods use the SVM with the RBF kernel as the classifier.
In addition, the compared HSI classification methods regard each training pixel as the positive instance for all the class labels contained in its candidate label set, as the previous HSI classification methods have never considered the partial label setting.

%We adopt the suggested settings in the original papers for all those baseline methods.
%We repeat the experiments for ten times with randomly sampled training pixels for all the methods and report the average results.
We use the recommended settings for all the compared methods in their papers. 
All the experiments are performed ten times with the random selection of the training pixels and the candidate labels, and the average results are presented.

\begin{figure}[t]
\subfigure[]{
\begin{minipage}[t]{\figsize\linewidth}
\centering
\includegraphics[width=1.0\textwidth , height=\heightfigure\textheight]{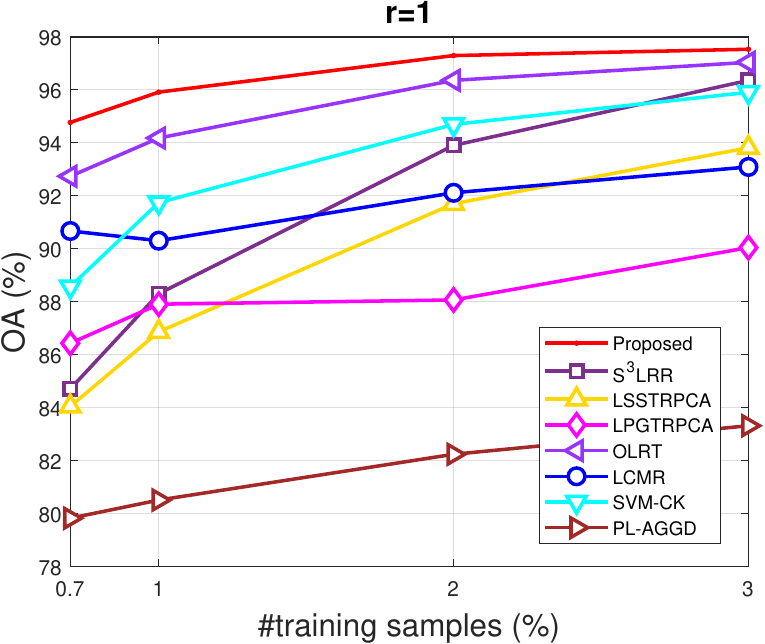}
\end{minipage}}
\centering
\subfigure[]{
\begin{minipage}[t]{\figsize\linewidth}
\centering
\includegraphics[width=1.0\textwidth , height=\heightfigure\textheight]{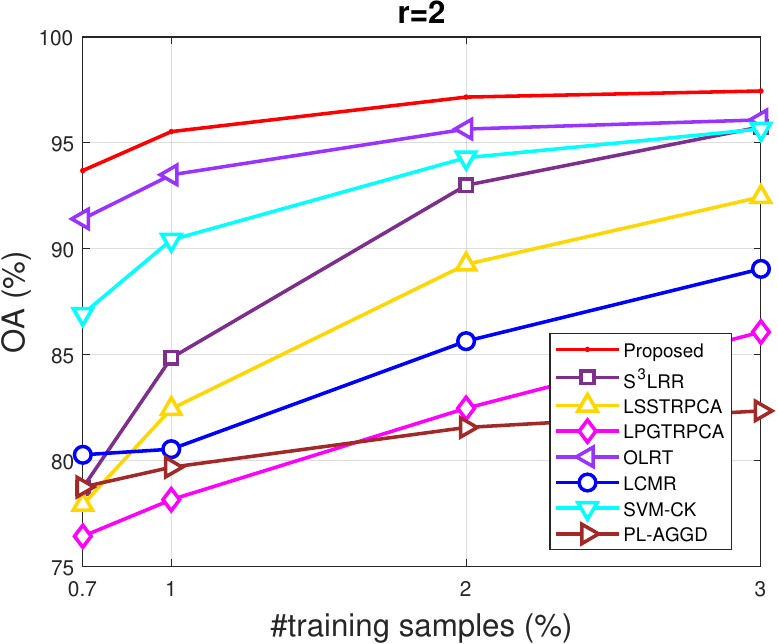}
\end{minipage}}

\subfigure[]{
\begin{minipage}[t]{\figsize\linewidth}
\centering
\includegraphics[width=1.0\textwidth , height=\heightfigure\textheight]{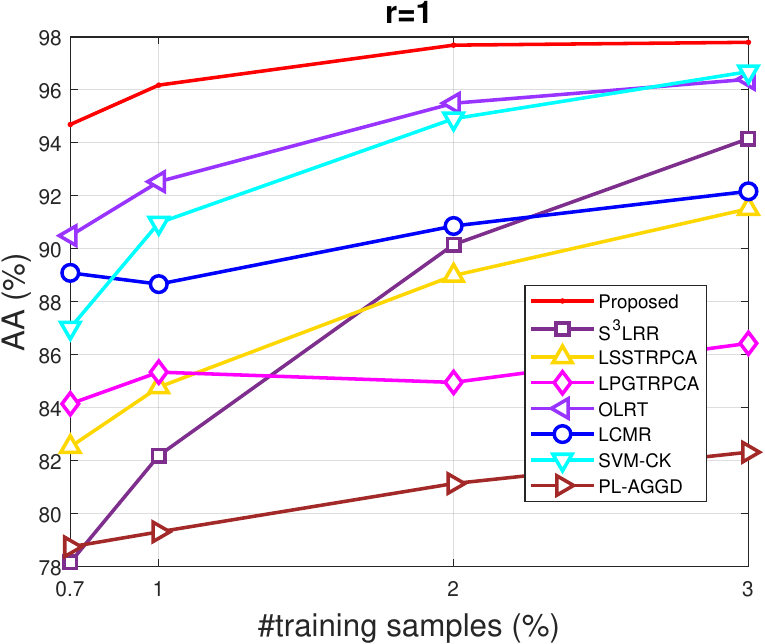}
\end{minipage}}
\centering
\subfigure[]{
\begin{minipage}[t]{\figsize\linewidth}
\centering
\includegraphics[width=1.0\textwidth , height=\heightfigure\textheight]{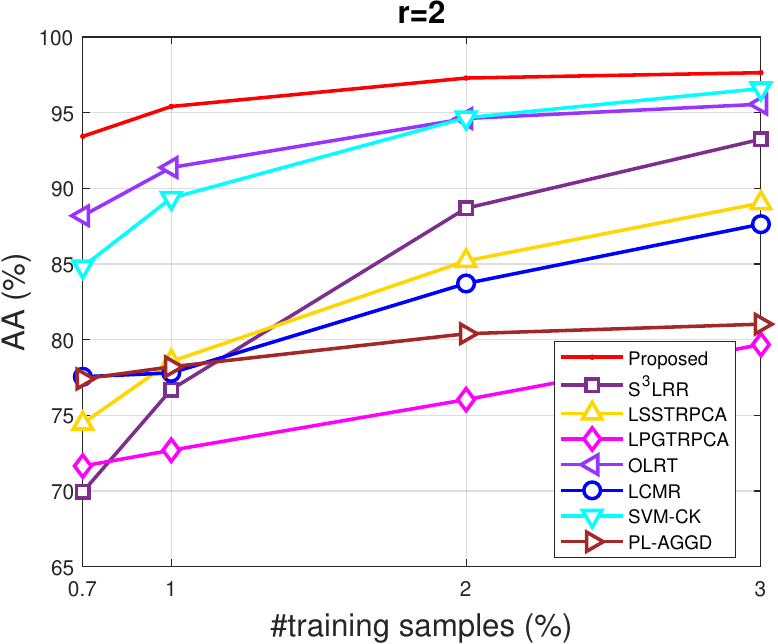}
\end{minipage}}

\subfigure[]{
\begin{minipage}[t]{\figsize\linewidth}
\centering
\includegraphics[width=1.0\textwidth , height=\heightfigure\textheight]{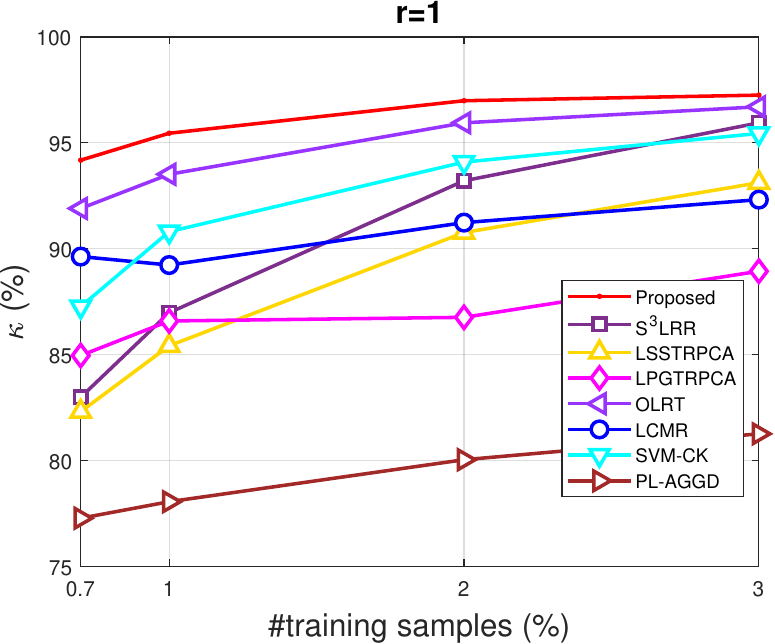}
\end{minipage}}
\centering
\subfigure[]{
\begin{minipage}[t]{\figsize\linewidth}
\centering
\includegraphics[width=1.0\textwidth , height=\heightfigure\textheight]{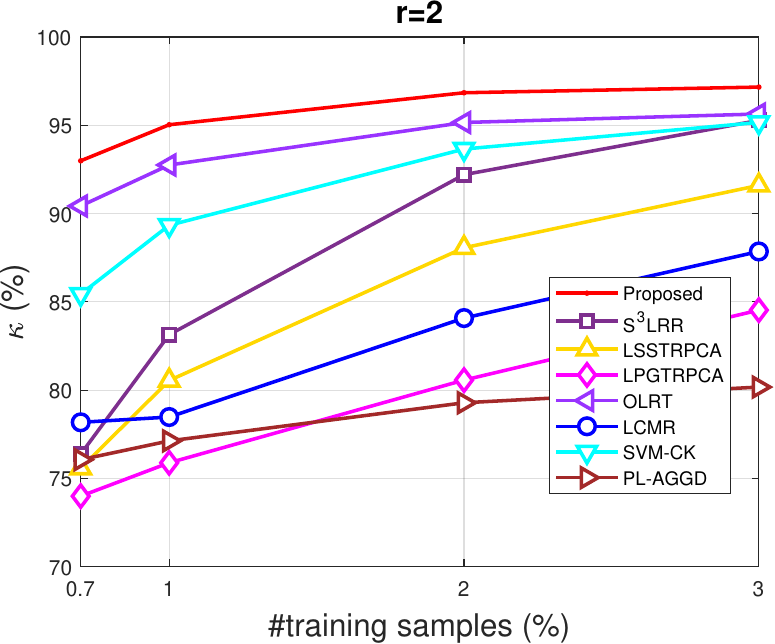}
\end{minipage}}
\caption{The performance comparisons concerning OA, AA, and $\kappa$ under various training percentages of each class with $r=1$ or 2 on \textit{Salinas Valley}. Specifically, (a)-(b) for OA with $r=1$ and 2, (c)-(d) for AA with $r=1$ and 2, and (e)-(f) for $\kappa$ with $r=1$ and 2. }
\label{fig:results_salinas}
\end{figure}

\begin{figure}[t]
\centering
\subfigure[]{
\begin{minipage}[t]{\figsizesmall\linewidth}
\centering
\includegraphics[width=1.0\textwidth , height=\heightfigures\textheight]{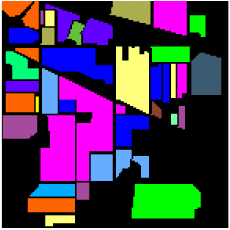}
\end{minipage}}
\centering
\subfigure[]{
\begin{minipage}[t]{\figsizesmall\linewidth}
\centering
\includegraphics[width=1.0\textwidth , height=\heightfigures\textheight]{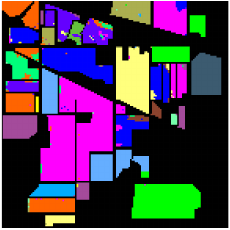}
\end{minipage}}
\centering
\subfigure[]{
\begin{minipage}[t]{\figsizesmall\linewidth}
\centering
\includegraphics[width=1.0\textwidth , height=\heightfigures\textheight]{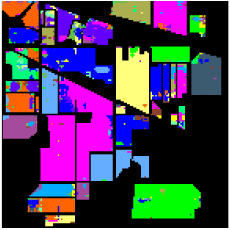}
\end{minipage}}
\centering
\subfigure[]{
\begin{minipage}[t]{\figsizesmall\linewidth}
\centering
\includegraphics[width=1.0\textwidth , height=\heightfigures\textheight]{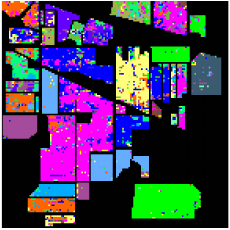}
\end{minipage}}
\centering
\subfigure[]{
\begin{minipage}[t]{\figsizesmall\linewidth}
\centering
\includegraphics[width=1.0\textwidth , height=\heightfigures\textheight]{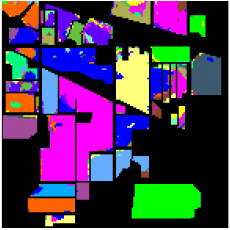}
\end{minipage}}

\centering
\subfigure[]{
\begin{minipage}[t]{\figsizesmall\linewidth}
\centering
\includegraphics[width=1.0\textwidth , height=\heightfigures\textheight]{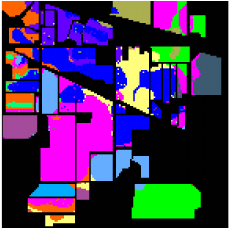}
\end{minipage}}
\centering
\subfigure[]{
\begin{minipage}[t]{\figsizesmall\linewidth}
\centering
\includegraphics[width=1.0\textwidth , height=\heightfigures\textheight]{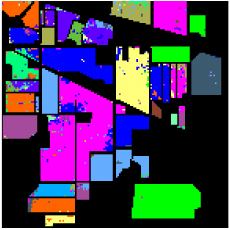}
\end{minipage}}
\centering
\subfigure[]{
\begin{minipage}[t]{\figsizesmall\linewidth}
\centering
\includegraphics[width=1.0\textwidth , height=\heightfigures\textheight]{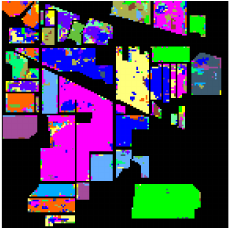}
\end{minipage}}
\centering
\subfigure[]{
\begin{minipage}[t]{\figsizesmall\linewidth}
\centering
\includegraphics[width=1.0\textwidth , height=\heightfigures\textheight]{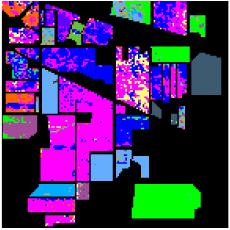}
\end{minipage}}
\centering
\caption{The classification maps of all the methods on \textit{Indian Pines} with 5\% training pixels per class and $r=1$. (a) Ground-truth; (b) Proposed (OA=93.69\%); (c) S$^3$LRR (OA=85.37\%); (d) LSSTRPCA (OA=74.14\%); (e) LPGTRPCA (OA=85.65\%); (f) OLRT (OA=74.27\%); (g) LCMR (OA=89.51\%); (h) SVM-CK (OA=76.74\%); (i) PL-AGGD (OA=65.52\%).}
\label{fig:map_indian}
\end{figure}

Figs. \ref{fig:results_indian} and \ref{fig:results_salinas} show the performance comparison between the proposed method and the compared methods under different training percentages per class with $r$=1 or 2 on \textit{Indian Pines} and \textit{Salinas Valley}, respectively. As illustrated by them, the proposed \proposed~method shows a remarkable superiority over all the compared methods with various training percentages. %Such an advantage of the proposed \proposed~method results from the fact that the previous methods, including the HSI classification methods and the classic partial label learning methods, never take into account both the partial label setting and the characteristic of the HSI data simultaneously. 
Such an advantage of the proposed \proposed~method comes from the lack of previous methods to combine the partial label setting and the data characteristics of an HSI.
%the previous HSI classification methods never consider the partial label learning and the previous partial label learning method do not take into account the characteristic of the HSI data.  
We can also observe that the performances of all methods gradually increase with the increasing training percentage. 
Furthermore, they almost decrease with $r$ increasing from 1 to 2 in most cases, indicating that more false labels (i.e., the bigger value of $r$) bring more classification difficulties. In addition, Fig. \ref{fig:map_indian} shows the classification maps of all the methods with $5\%$ training pixels per class and $r=1$ on \textit{Indian Pines}, further demonstrating the superiority of the proposed method.
%\subsection{Comparison of Running Time}
\section{Conclusion}\label{sec:conclusion}
In this letter, we propose the \proposed~method, the first concerning partial label learning problem in HSI classification. Owing to the exploration of discriminative representations and disambiguated training labels, our \proposed~method can alleviate both the feature ambiguity (i.e., spectral variations) and the label ambiguity (i.e., partial label learning).
%Specifically, our \proposed~method proposes a novel superpixelwise LRA-based model, which is capable of extracting the discriminative representations while preparing the affinity graph for the subsequent learning process. After that, it employs a label propagation procedure to disambiguate the training labels. The resulting discriminative representations and the disambiguated labels are further used to boost the classification results. 
The experiments demonstrate the advantage of the proposed \proposed~method compared with the state-of-the-art methods.
%\balance
%\bibliographystyle{ieeetr}
%\footnotesize
\bibliographystyle{IEEEtran}
\bibliography{pl}

\begin{thebibliography}{10}
\providecommand{\url}[1]{#1}
\csname url@samestyle\endcsname
\providecommand{\newblock}{\relax}
\providecommand{\bibinfo}[2]{#2}
\providecommand{\BIBentrySTDinterwordspacing}{\spaceskip=0pt\relax}
\providecommand{\BIBentryALTinterwordstretchfactor}{4}
\providecommand{\BIBentryALTinterwordspacing}{\spaceskip=\fontdimen2\font plus
\BIBentryALTinterwordstretchfactor\fontdimen3\font minus \fontdimen4\font\relax}
\providecommand{\BIBforeignlanguage}[2]{{%
\expandafter\ifx\csname l@#1\endcsname\relax
\typeout{** WARNING: IEEEtran.bst: No hyphenation pattern has been}%
\typeout{** loaded for the language `#1'. Using the pattern for}%
\typeout{** the default language instead.}%
\else
\language=\csname l@#1\endcsname
\fi
#2}}
\providecommand{\BIBdecl}{\relax}
\BIBdecl

\bibitem{mei2018simultaneous}
S.~Mei, J.~Hou, J.~Chen, L.-P. Chau, and Q.~Du, ``Simultaneous spatial and spectral low-rank representation of hyperspectral images for classification,'' \emph{IEEE Transactions on Geoscience and Remote Sensing}, vol.~56, no.~5, pp. 2872--2886, 2018.

\bibitem{sun2019lateral}
W.~Sun, G.~Yang, J.~Peng, and Q.~Du, ``Lateral-slice sparse tensor robust principal component analysis for hyperspectral image classification,'' \emph{IEEE Geoscience and Remote Sensing Letters}, vol.~17, no.~1, pp. 107--111, 2020.

\bibitem{wang2021tensor}
Y.~Wang, T.~Li, L.~Chen, Y.~Yu, Y.~Zhao, and J.~Zhou, ``Tensor-based robust principal component analysis with locality preserving graph and frontal slice sparsity for hyperspectral image classification,'' \emph{IEEE Transactions on Geoscience and Remote Sensing}, 2021.

\bibitem{chang2020hyperspectral}
Y.~Chang, L.~Yan, B.~Chen, S.~Zhong, and Y.~Tian, ``Hyperspectral image restoration: Where does the low-rank property exist,'' \emph{IEEE Transactions on Geoscience and Remote Sensing}, vol.~59, no.~8, pp. 6869--6884, 2021.

\bibitem{liu2019review}
S.~Liu, D.~Marinelli, L.~Bruzzone, and F.~Bovolo, ``A review of change detection in multitemporal hyperspectral images: Current techniques, applications, and challenges,'' \emph{IEEE Geoscience and Remote Sensing Magazine}, vol.~7, no.~2, pp. 140--158, 2019.

\bibitem{zhang2020improved}
W.~Zhang, P.~Du, C.~Lin, P.~Fu, X.~Wang, X.~Bai, H.~Zheng, J.~Xia, and A.~Samat, ``An improved feature set for hyperspectral image classification: Harmonic analysis optimized by multiscale guided filter,'' \emph{IEEE Journal of Selected Topics in Applied Earth Observations and Remote Sensing}, vol.~13, pp. 3903--3916, 2020.

\bibitem{kang2013feature}
X.~Kang, S.~Li, and J.~A. Benediktsson, ``Feature extraction of hyperspectral images with image fusion and recursive filtering,'' \emph{IEEE Transactions on Geoscience and Remote Sensing}, vol.~52, no.~6, pp. 3742--3752, 2013.

\bibitem{xu2021dual}
Y.~Xu, Z.~Li, W.~Li, Q.~Du, C.~Liu, Z.~Fang, and L.~Zhai, ``Dual-channel residual network for hyperspectral image classification with noisy labels,'' \emph{IEEE Transactions on Geoscience and Remote Sensing}, vol.~60, pp. 1--11, 2021.

\bibitem{tu2020hyperspectral}
B.~Tu, C.~Zhou, D.~He, S.~Huang, and A.~Plaza, ``Hyperspectral classification with noisy label detection via superpixel-to-pixel weighting distance,'' \emph{IEEE Transactions on Geoscience and Remote Sensing}, vol.~58, no.~6, pp. 4116--4131, 2020.

\bibitem{wang2019partial}
Q.-W. Wang, Y.-F. Li, and Z.-H. Zhou, ``Partial label learning with unlabeled data.'' in \emph{IJCAI}, 2019, pp. 3755--3761.

\bibitem{wang2021adaptive}
D.-B. Wang, M.-L. Zhang, and L.~Li, ``Adaptive graph guided disambiguation for partial label learning,'' \emph{IEEE Transactions on Pattern Analysis \& Machine Intelligence}, vol.~44, no.~12, pp. 8796--8811, 2022.

\bibitem{zhang2015solving}
M.-L. Zhang and F.~Yu, ``Solving the partial label learning problem: An instance-based approach.'' in \emph{IJCAI}, 2015, pp. 4048--4054.

\bibitem{xu2015spectral}
Y.~Xu, Z.~Wu, and Z.~Wei, ``Spectral--spatial classification of hyperspectral image based on low-rank decomposition,'' \emph{IEEE Journal of Selected Topics in Applied Earth Observations and Remote Sensing}, vol.~8, no.~6, pp. 2370--2380, 2015.

\bibitem{fan2017hyperspectral}
F.~Fan, Y.~Ma, C.~Li, X.~Mei, J.~Huang, and J.~Ma, ``Hyperspectral image denoising with superpixel segmentation and low-rank representation,'' \emph{Information Sciences}, vol. 397, pp. 48--68, 2017.

\bibitem{liu2011entropy}
M.-Y. Liu, O.~Tuzel, S.~Ramalingam, and R.~Chellappa, ``Entropy rate superpixel segmentation,'' in \emph{CVPR}, 2011, pp. 2097--2104.

\bibitem{yang2022local}
S.~Yang, Y.~Zhang, Y.~Jia, and W.~Zhang, ``Local low-rank approximation with superpixel-guided locality preserving graph for hyperspectral image classification,'' \emph{IEEE Journal of Selected Topics in Applied Earth Observations and Remote Sensing}, vol.~15, pp. 7741--7754, 2022.

\bibitem{lin2010augmented}
Z.~Lin, M.~Chen, and Y.~Ma, ``The augmented lagrange multiplier method for exact recovery of corrupted low-rank matrices,'' \emph{Unpublished paper}, 2010, [Online]. Available:\url{https://arxiv.org/abs/1009.5055}.

\bibitem{liu2010robust}
G.~Liu, Z.~Lin, and Y.~Yu, ``Robust subspace segmentation by low-rank representation,'' in \emph{Proceedings of the 27th international conference on machine learning (ICML-10)}, 2010, pp. 663--670.

\bibitem{lcmr}
L.~Fang, N.~He, S.~Li, A.~J. Plaza, and J.~Plaza, ``A new spatial–spectral feature extraction method for hyperspectral images using local covariance matrix representation,'' \emph{IEEE Transactions on Geoscience and Remote Sensing}, vol.~56, no.~6, pp. 3534--3546, 2018.

\bibitem{svm-ck}
G.~Camps-Valls, L.~Gómez-Chova, J.~Muñoz-Marí, J.~Vila-Francés, and J.~Calpe-Maravilla, ``Composite kernels for hyperspectral image classification.'' \emph{IEEE Geoscience and Remote Sensing Letters}, vol.~3, no.~1, pp. 93--97, 2006.

\end{thebibliography}
%\appendix
\end{document}